\documentclass[10pt,twocolumn,letterpaper]{article}

\usepackage[pagenumbers]{cvpr}

\usepackage{booktabs}
\usepackage{graphicx}
\usepackage{multirow}
\usepackage{xcolor}
\usepackage{colortbl}
\usepackage{tabularx}
\usepackage{float}

\definecolor{groupgray}{RGB}{245,246,248}
\definecolor{oursblue}{RGB}{230,240,247}
\definecolor{origgray}{gray}{0.45}
\providecommand{\orderscore}[2]{}
\renewcommand{\orderscore}[2]{%
  \mbox{%
    \makebox[2.60em][r]{#1}\hspace{0.12em}%
    \raisebox{-0.30ex}{%
      \scriptsize\color{origgray}%
      \makebox[3.10em][r]{\ensuremath{#2}}%
    }%
  }%
}

\definecolor{cvprblue}{rgb}{0.21,0.49,0.74}
\usepackage[
    pagebackref,
    breaklinks,
    colorlinks,
    allcolors=cvprblue
]{hyperref}

\def\confName{CVPR}
\def\confYear{2027}

\title{SlideJev: Order-Stable and Tissue-Grounded Jev Decisions\\
for Whole-Slide Pathology}

\author{
Lei Wu$^{1,*}$,
Jiashuai Liu$^{1,*}$,
Di Zhang$^{1}$,
Chunze Yang$^{1}$,
Jiusong Ge$^{1}$,
Junbo Lu$^{1}$\\
Yongqi Liang$^{1}$,
Wenjie Zhao$^{1}$,
Ni Zhang$^{1}$,
Chen Li$^{2,\dagger}$,
Zeyu Gao$^{3,\dagger}$,
Qidong Liu$^{1,\dagger}$\\[0.6em]
{\small
$^{1}$School of Computer Science and Technology,
Xi'an Jiaotong University}\\
{\small
$^{2}$Institute of Intelligent Medicine,
Chinese Academy of Medical Sciences}\\
{\small
$^{3}$University of Cambridge}
}

\begin{document}

\maketitle

\makeatletter
\begingroup
  \renewcommand{\thefootnote}{\fnsymbol{footnote}}
  \long\def\@makefntext#1{%
    \parindent=0pt
    \leftskip=0pt
    \rightskip=0pt plus 1fil
    \parfillskip=0pt
    \noindent\@makefnmark\,#1\par
  }
  \footnotetext[1]{These authors contributed equally.}
  \footnotetext[2]{%
    Corresponding authors:\newline
    Chen Li: \texttt{cli@pumc.edu.cn}\newline
    Zeyu Gao: \texttt{zg323@cam.ac.uk}\newline
    Qidong Liu: \texttt{liuqidong@xjtu.edu.cn}%
  }
\endgroup
\makeatother

\suppressfloats[t]

\begin{abstract}
Whole-slide visual question answering (VQA) offers a
natural-language interface to pathology, allowing diagnostic
and phenotypic questions to be answered directly from
tissue slides.
However, current vision-language models (VLMs) rely heavily
on linguistic priors: their answers are sensitive to the
order of the candidate answers, and can be reached through
shortcuts that bypass the tissue evidence.
To address these challenges, we introduce \textbf{SlideJev},
a framework that separately verifies candidate answers
against the slide and learns to ground these judgments
in tissue evidence.
To connect parallel candidate evaluation with visual
evidence learning, we adopt Jev and score each candidate
by the difference between support and rejection
log-likelihoods from the original language-model head.
We compute these scores in parallel using separate prompts
with the same format and no option-position labels,
then normalize them for answer selection.
To ground these scores in tissue evidence,
candidate-conditioned reading and comparison guide gated
updates to visual tokens across decoder layers.
With the backbone frozen, we train these visual modules
through the same scores using question-answer and
paired-image supervision.
Across three whole-slide pathology VQA benchmarks,
SlideJev achieves the highest mean overall accuracy and
100\% answer consistency across four option orders.
Image-control experiments further show that SlideJev
achieves the largest average accuracy gains from correct
images over mismatched and image-free inputs across
the three datasets.
Our code is publicly available on
\href{https://github.com/leowu-ai/SlideJev}{GitHub}.
\end{abstract}
\section{Introduction}
\label{sec:introduction}

Pathological decision-making relies on interpreting cellular
morphology and tissue architecture in hematoxylin and eosin
(H\&E) whole-slide images (WSIs) to support diagnosis and
phenotypic assessment~\cite{lu2024visual}.
Whole-slide visual question answering (VQA) provides a
natural-language interface to this process, predicting
an answer from a WSI and a pathology-related
question~\cite{chen2025slidechat,liang2025wsi}.
A common form is multiple-choice VQA, where the model
selects among candidate diagnostic or phenotypic
interpretations~\cite{lu2024multimodal}.
However, pathology questions span diverse medical concepts,
and the relevant visual findings vary with what is being asked.
Answering them requires understanding medical language,
recognizing morphological patterns, and relating these
patterns to the diagnostic or phenotypic concepts expressed
in the question.

\begin{figure}[t]
    \centering
    \includegraphics[width=\columnwidth]
    {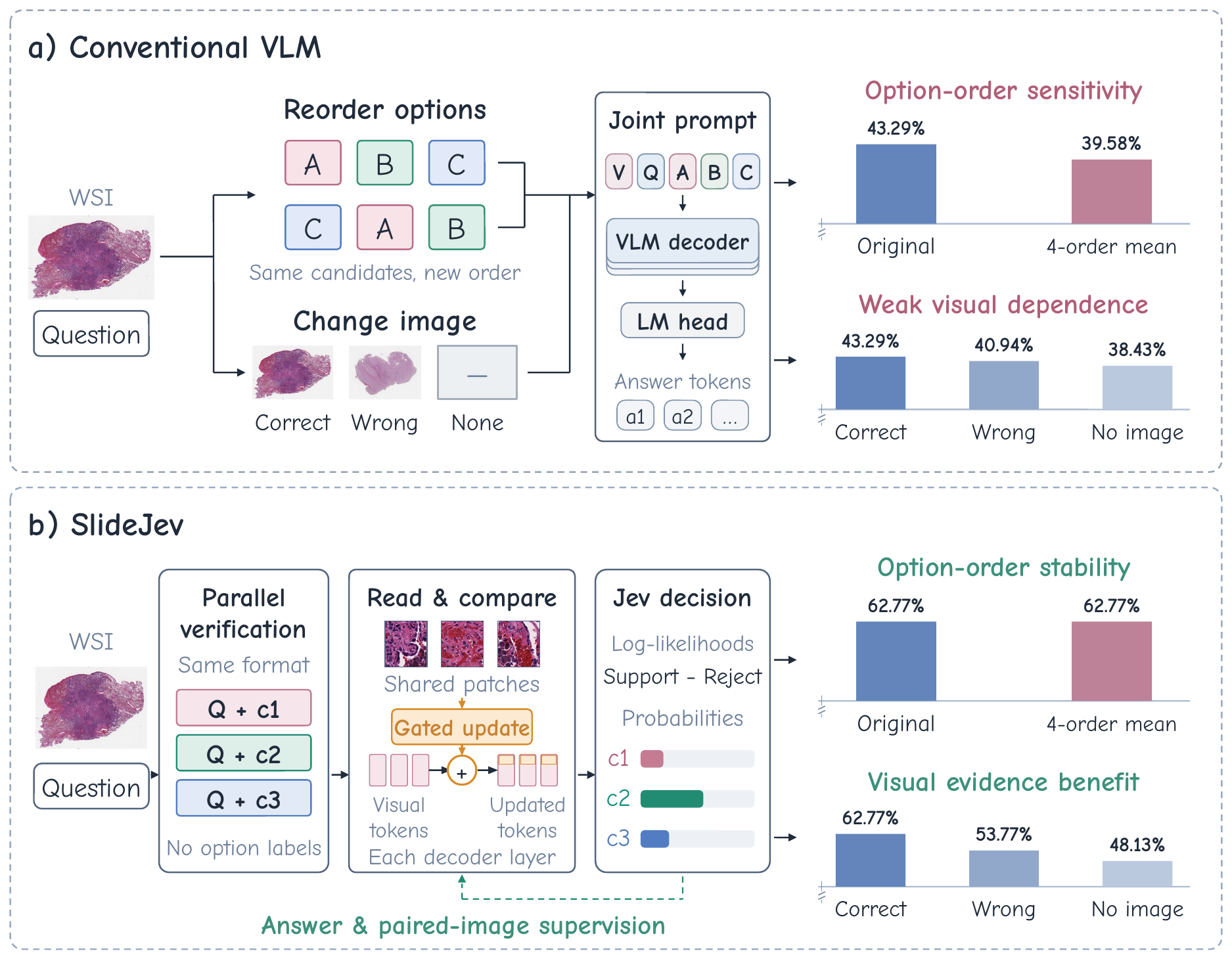}
\caption{\textbf{Reliability challenges and SlideJev.}
(a) Order sensitivity and weak visual dependence.
(b) Jev links parallel verification and gated updates
for stable, evidence-grounded decisions.}
    \label{fig:intro}
    \vspace{-2mm}
\end{figure}

These requirements motivate the use of vision-language
models (VLMs), which connect visual representations
with language understanding for pathology question
answering~\cite{lu2024multimodal,seyfioglu2024quilt}.
When applied to WSIs through thumbnails or selected tissue
images, \textit{Image-level VLMs} face limitations in
morphological detail or tissue coverage~\cite{chen2025slidechat}.
By comparison, \textit{Whole-slide VLMs} incorporate tissue representations
from across the slide, providing access to information
beyond individual image fields~\cite{shaikovski2024prism,
ding2025multimodal}.
They use the visual context alongside the question
and candidate descriptions to predict an answer.
Despite richer visual context, \textbf{Conventional VLM} (shown in Fig~\ref{fig:intro}(a)) still face
two reliability challenges when evaluating candidate
answers in multiple-choice VQA.
\textbf{i) Option-order Sensitivity.}
When candidates are presented jointly, their positions
become part of the prompt context~\cite{zheng2023large,
pezeshkpour2024large}.
Reordering them can change the selected answer even
when the question and slide remain unchanged.
\textbf{ii) Visual Evidence Underutilization.}
Access to whole-slide information does not ensure that
candidate judgments respond to relevant tissue
findings~\cite{liang2025wsi,goyal2017making}.
Prediction accuracy may change little when the slide
is replaced with a mismatched one or removed.

Reducing the option-order sensitivity identified above calls
for evaluating candidates separately under a consistent
format, while strengthening visual evidence use calls
for their assessments to be refined by discriminative
tissue information.
Jev exposes a probability distribution over a predefined
answer space, making the relative support for competing
answers explicit~\cite{typesafe2026jev,li2026llmasjev}.
This formulation provides a useful basis for expressing
and comparing candidate preferences.
Adapting it to whole-slide VQA, however, requires a common
scoring rule for separate candidate evaluations and
a trainable connection between tissue evidence and
the resulting scores.
The same scores should support direct answer selection
and provide supervision that encourages candidate
preferences to change with relevant tissue findings,
connecting separate verification with the learning
of visual evidence updates.

To support stable pathological decisions grounded in
tissue evidence, we introduce \textbf{SlideJev}, a pathology
VQA framework following the whole-slide VLM paradigm,
as illustrated in \cref{fig:intro}(b).
Overall, we construct a \textbf{Jev Decision Interface}
with comparable candidate scores for answer selection
and visual evidence learning.
To reduce option-order sensitivity, we verify candidates
separately through \textbf{Parallel Candidate Inputs}
in a shared prompt format without option-position labels,
enabling order-stable selection through Jev.
To ground these assessments in tissue evidence, we then
introduce \textbf{Gated Visual Injection}, using
candidate-relevant evidence and response comparisons
on shared regions to guide visual updates.
The same Jev scores supervise these updates, encouraging
candidate preferences to respond to tissue evidence.
We evaluate SlideJev on the TCGA subset of
SlideBench-VQA~\cite{chen2025slidechat}
and assess transfer on BCNB~\cite{xu2021predicting}
and BRACS-3~\cite{brancati2022bracs}.
SlideJev achieves the highest four-order mean overall
accuracy among the compared methods on all three benchmarks.
Our contributions are summarized as follows:
\begin{itemize}
    \item We present SlideJev, a whole-slide pathology VQA
    framework that uses Jev as a shared basis for candidate
    selection and visual evidence learning.

    \item We formulate parallel candidate inputs that preserve
    verification contexts under option reordering and
    combine separate assessments through a common Jev
    readout for order-stable answer selection.

    \item We develop gated visual injection that integrates
discriminative tissue evidence with Jev-based supervision
to strengthen image-dependent candidate judgments.
\end{itemize}
\section{Preliminaries}
\label{sec:preliminaries}

\subsection{Whole-slide VLM pipeline}

A typical \textbf{generative whole-slide VLM pipeline}
comprises visual representation, multimodal processing,
and response generation~\cite{chen2025slidechat,liang2025wsi}.
Given a whole-slide image $I$, a question $Q$, and candidate
answers $C=\{c_i\}_{i=1}^{K}$ with $K\ge2$, the model
obtains slide representations and provides them to a language
decoder together with the question and candidate descriptions.
Let $x$ denote this multimodal context and $a$ the response
token sequence.
To predict response token $a_t$, the decoder computes
a final-layer state $h_t$ that incorporates information
from $x$ and the preceding response tokens $a_{<t}$.
The language-model head assigns a probability to $a_t$ as
\begin{equation}
P_{\theta}(a_t\mid x,a_{<t})
=
\left[
\operatorname{softmax}(W_{\mathrm{LM}}h_t)
\right]_{a_t},
\label{eq:vlm_generation}
\end{equation}
where $\theta$ denotes model parameters and
$W_{\mathrm{LM}}$ maps decoder states to vocabulary logits.
The response is generated autoregressively and identifies
the selected candidate through an option label or answer text.
\textit{Within this pipeline, SlideJev refines candidate input
organization, visual evidence processing during decoding,
and answer readout to support order-stable decisions
grounded in tissue evidence}.

\subsection{Language Decision Model: Jev}

\textbf{Jev} uses a decision-oriented formulation with
an answer space specified by the question and candidate
descriptions~\cite{typesafe2026jev,li2026llmasjev}.
For choice tasks, it returns a probability distribution
over the supplied candidates and selects the most probable one.
Its output makes the relative support for competing answers
explicit at the candidate level, while the generative
formulation above predicts vocabulary tokens that compose
a response.
\section{Method}
\label{sec:method}

\begin{figure*}[t]
    \centering
    \includegraphics[width=\textwidth]{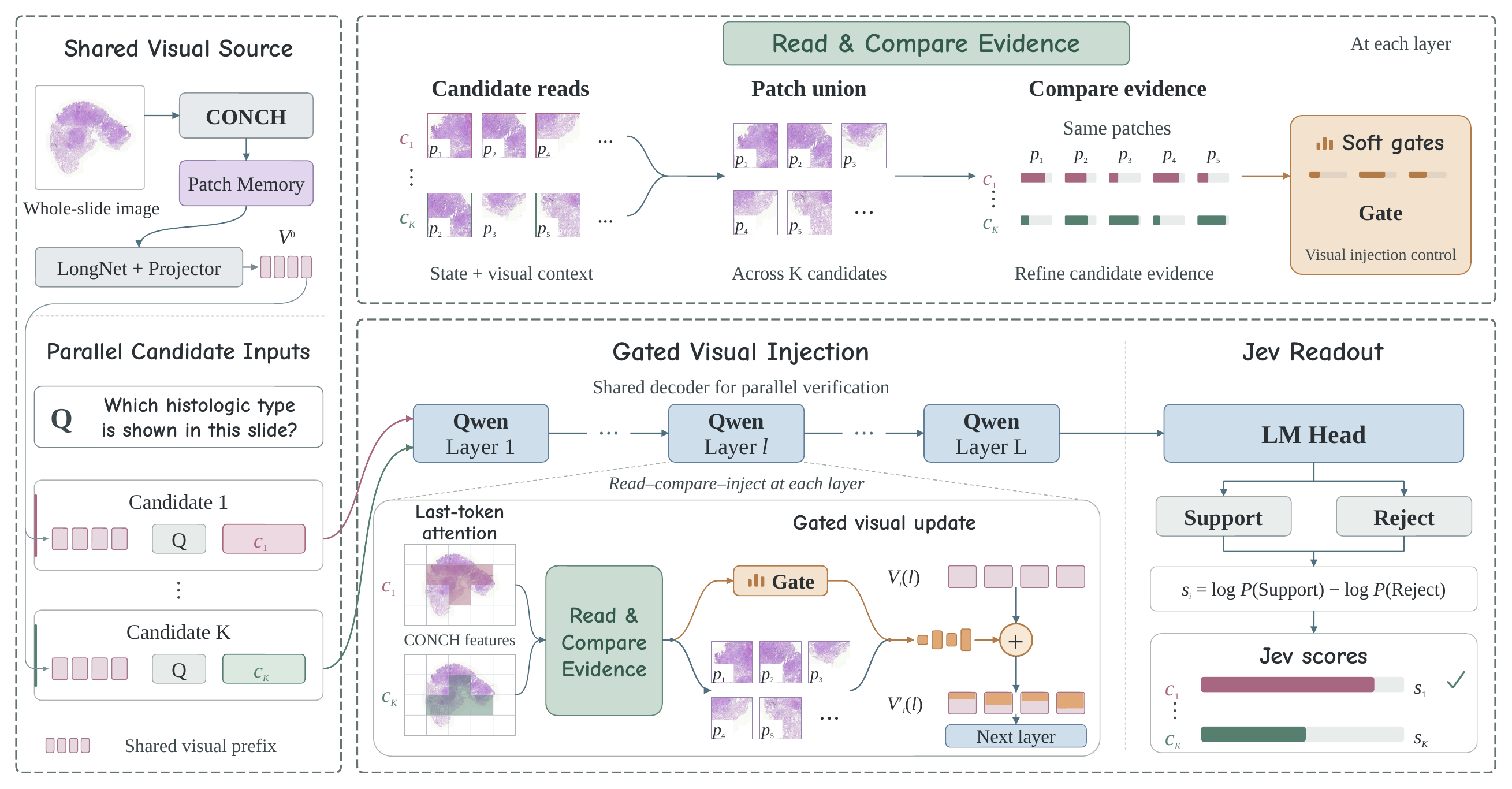}
    \caption{Overview of \textbf{SlideJev}.
    A shared visual source supplies CONCH patch memory and an initial visual prefix.
    Candidates are verified in parallel using separate, identically formatted prompts; only the first and last of $K$ inputs are shown.
    At each decoder layer, candidate-conditioned reads retrieve patch evidence, and candidate responses are compared on the union of selected patches.
    The enlarged inset illustrates how soft gates control visual-token updates before each frozen Qwen decoder block.
    The Jev readout contrasts support and rejection log-likelihoods from the original language-model head to produce candidate scores.
    }
    \label{fig:overview}
\end{figure*}

We present \textbf{SlideJev}, a whole-slide pathology VQA
framework that connects independent candidate verification
with visual evidence learning, as illustrated in
\cref{fig:overview}.
Following the whole-slide VLM paradigm, we organize the
decision pipeline around parallel candidate inputs,
gated visual injection, and a Jev readout.
First, we construct \textbf{Parallel Candidate Inputs}
by pairing each candidate with the same question and
shared visual prefix without option-position labels,
preserving its verification context under option reordering.
Next, candidate-conditioned reads retrieve evidence from
the shared patch memory, while cross-candidate comparison
evaluates differences in responses to shared tissue patches.
The retrieved evidence and comparison signals are
incorporated through \textbf{Gated Visual Injection}
across decoder layers to refine candidate assessment.
Finally, the \textbf{Jev Readout} contrasts support and
rejection log-likelihoods from the original language-model
head, yielding candidate scores that are jointly normalized
for answer selection.
During training, question-answer and paired-image supervision
optimize the visual modules through the same scores,
encouraging candidate preferences to reflect image content
while keeping the pretrained backbone fixed.

\subsection{Parallel Candidate Inputs}
\label{sec:parallel_verification}

We construct parallel candidate inputs to keep each
verification prompt unchanged under option reordering.
For slide $I$, we use SlideChat's frozen visual pathway
to obtain a shared patch memory $F=[f_n]_{n=1}^{N}$
and an initial visual prefix $V^0$, where $N$ is the
number of patches.

Each candidate is paired with $Q$ in a binary verification
prompt asking whether it correctly answers the question
based on $I$.
All prompts share the same format and omit the original
option-position labels.
The inputs are constructed as
\begin{equation}
x_i=(V^0,Q,c_i),
\qquad i=1,\ldots,K.
\label{eq:parallel_inputs}
\end{equation}
The shared frozen decoder processes these inputs in
parallel, keeping their text sequences separate.
The initially shared visual tokens undergo candidate-specific
updates, with cross-candidate interaction provided by
the evidence comparison module described next.
 
\subsection{Evidence Reading And Comparison}
\label{sec:evidence_reading}

This module extracts discriminative tissue information
for candidate verification, producing candidate-specific
evidence and pairwise comparison messages for visual injection.
At decoder layer $l$, let $h_i(l)$ denote the hidden state at the final verification-prompt position and $V_i(l)$ the current visual tokens for candidate $c_i$, initialized as $V_i(1)=V^0$.

\vspace{1mm}
\noindent\textbf{Evidence Reading.}
We use the candidate's current state and visual context to retrieve evidence from the full patch memory.
A query $q_i(l)=W_q\operatorname{LN}(h_i(l))$ first reads the current visual tokens to obtain a context summary $t_i(l)=\operatorname{Attn}(q_i(l),V_i(l))$.
This summary refines the query as $\tilde q_i(l)=q_i(l)+W_c t_i(l)$.
With patch keys $k_n=W_k\operatorname{LN}(f_n)$,
we retrieve evidence from $F$:
\begin{equation}
e_i(l)
=
\sum_{n=1}^{N}
\frac{
\exp(\tilde q_i(l)^{\top}k_n/\sqrt d)
}{
\sum_{r=1}^{N}\exp(\tilde q_i(l)^{\top}k_r/\sqrt d)
}
f_n.
\label{eq:evidence_reading}
\end{equation}
Here, $W_q$, $W_k$, and $W_c$ are learned projections,
and $d$ is the query and key dimension.
The normalized coefficients form an attention vector
$\alpha_i(l)$ over the $N$ patches.
The original patch features serve directly as values.

\vspace{1mm}
\noindent\textbf{Shared Patch Selection.}
We establish a common spatial basis for comparison,
including patches strongly attended by different candidates.
Let $R_i(l)$ contain the patch indices corresponding to
the top-$T$ entries of $\alpha_i(l)$.
We form the shared patch set
$R_l=\bigcup_{i=1}^{K}R_i(l)$.
If the union exceeds the comparison budget, we retain
patches with the largest summed attention across candidates.
This budget restricts comparison, while evidence retrieval
continues to use the full memory.
The retained patch features form the columns of
$F_{R_l}=[f_n]_{n\in R_l}$.

\vspace{1mm}
\noindent\textbf{Candidate Matching.}
Every candidate is rescored on every retained patch.
The matching scores form the vector
$a_i(l)=[\phi_q(\tilde q_i(l))^\top
\phi_k(k_n)/\sqrt d]_{n\in R_l}$,
where $\phi_q$ and $\phi_k$ are learned matching projections.
For each candidate pair $(i,j)$ and retained patch,
a shared network combines the patch feature with symmetric
sums and absolute differences of the candidate queries,
patch-level matching scores, and retrieval attention weights.
Its outputs are normalized over $R_l$ to form the
patch-weight vector $w_{ij}(l)$.
All patch-level vectors follow the column order of $F_{R_l}$.

\vspace{1mm}
\noindent\textbf{Evidence Comparison.}
We aggregate patch-level differences into a comparison message
for each candidate pair:
\begin{equation}
m_{ij}(l)
=
F_{R_l}\cdot
\Bigl(
w_{ij}(l)\odot
\tanh\bigl(a_i(l)-a_j(l)\bigr)
\Bigr).
\label{eq:evidence_comparison}
\end{equation}
Here, $\odot$ denotes elementwise multiplication,
and $\tanh$ is applied elementwise.
The weights control each patch's contribution, while
the signed score differences indicate which candidate
has the higher patch-level matching score.
Symmetric pair features give $w_{ij}(l)=w_{ji}(l)$,
so exchanging the candidates reverses the message:
$m_{ji}(l)=-m_{ij}(l)$.
The retrieved evidence $e_i(l)$ and comparison messages
$m_{ij}(l)$ then condition gated visual injection.

\subsection{Gated Visual Injection}
\label{sec:visual_injection}

This module converts evidence and comparison
messages into candidate-specific visual updates, with
learned gates controlling their strength across decoder layers.

\vspace{1mm}
\noindent\textbf{Evidence Routing and Fusion.}
Before evidence retrieval at layer $l$ of an $L$-block decoder,
a shared router predicts symmetric pair gates
$z_{ij}(l)=z_{ji}(l)\in(0,1)$ from the current queries
$q_i(l)$ and $q_j(l)$, evidence cached from preceding layers,
cache-validity indicators, and normalized depth $l/L$.
The validity indicators specify which cached evidence
is available to the router.
The pair gates are aggregated into a candidate injection gate
$g_i(l)=1-\prod_{j\neq i}(1-z_{ij}(l))$.
After each read, the retrieved evidence is cached
for subsequent layers.
Retrieved evidence and gated comparison messages jointly
form a visual update condition:
\begin{equation}
b_i(l)
=
W_e e_i(l)
+
\frac{W_m}{K-1}
\sum_{j\neq i}z_{ij}(l)m_{ij}(l).
\label{eq:evidence_fusion}
\end{equation}
Here, $W_e$ and $W_m$ project the evidence and comparison
messages into the decoder hidden space. The first term
supplies the candidate's evidence, while the second
averages its gated comparison messages over the other candidates.

\vspace{1mm}
\noindent\textbf{Gated Visual Update.}
A shared MLP combines each visual token's current state
with the fused condition to generate token-specific residuals:
\begin{equation}
D_i(l)
=
\operatorname{MLP}\bigl(V_i(l),b_i(l)\bigr).
\label{eq:visual_residual}
\end{equation}
The MLP operates independently at each visual position,
with $b_i(l)$ broadcast across positions.
The resulting residual matrix $D_i(l)$ has the same shape
as $V_i(l)$.
The candidate gate then controls the visual update:
\begin{equation}
V'_i(l)
=
V_i(l)+\eta_l g_i(l)D_i(l),
\label{eq:visual_write}
\end{equation}
where $\eta_l$ is a learned bounded gain for layer $l$.
The updated prefix enters the frozen decoder block
with the original causal mask, and the block outputs
provide the states for the next layer.
At inference, soft gates are computed at every decoder layer.
The added modules share parameters across layers,
except for the layer-specific gains $\eta_l$.

\begin{table*}[t]
\centering
\caption{\textbf{Results on SlideBench-VQA (TCGA) and BRACS-3.}
Each cell reports the four-order mean (\%); smaller gray numbers
indicate the difference from the original-order result in
percentage points (pp).
TCGA columns report accuracy, as answer choices vary by question.
For BRACS-3 transfer, ACC reports accuracy and BA denotes
balanced accuracy, computed as the mean recall across
the three classes.}
\label{tab:tcga_bracs}

\small
\setlength{\tabcolsep}{2.5pt}
\renewcommand{\arraystretch}{1.10}

\begin{tabularx}{\textwidth}{
    @{}l*{6}{>{\centering\arraybackslash}X}@{}
}
\toprule[0.8pt]
\multirow{2}{*}{\textbf{Model}}
& \multicolumn{4}{c}{\textbf{TCGA}}
& \multicolumn{2}{c}{\textbf{BRACS-3}} \\
\cmidrule(lr){2-5}
\cmidrule(lr){6-7}
& Microscopy & Diagnosis & Clinical & \textbf{Overall}
& \textbf{ACC} & \textbf{BA} \\
\midrule

\rowcolor{groupgray}
\multicolumn{7}{l}{\textbf{(a) General-purpose VLMs}} \\

Gemma-4-E4B-it
& \orderscore{48.94}{-5.41}
& \orderscore{42.02}{-4.28}
& \orderscore{60.67}{-5.57}
& \orderscore{44.99}{-4.67}
& \orderscore{49.08}{-0.74}
& \orderscore{41.63}{-0.72} \\

Qwen3.6-27B
& \orderscore{63.18}{-2.58}
& \orderscore{40.58}{-3.97}
& \orderscore{71.18}{-2.71}
& \orderscore{48.75}{-3.50}
& \orderscore{53.91}{+0.97}
& \orderscore{39.31}{+1.35} \\

Qwen3.8-27B
& \orderscore{65.57}{-3.13}
& \orderscore{46.26}{-0.72}
& \orderscore{71.97}{-1.28}
& \orderscore{53.23}{-1.45}
& \orderscore{50.78}{-0.51}
& \orderscore{36.26}{-1.38} \\

GPT-6.1-sol
& \orderscore{69.40}{-1.36}
& \orderscore{53.24}{-0.61}
& \orderscore{71.02}{+0.96}
& \orderscore{58.88}{-0.75}
& \orderscore{54.69}{+1.93}
& \orderscore{\underline{43.30}}{+1.74} \\
\midrule

\rowcolor{groupgray}
\multicolumn{7}{l}{\textbf{(b) Medical and pathology VLMs}} \\

Quilt-LLaVA
& \orderscore{46.30}{-0.87}
& \orderscore{32.30}{+0.80}
& \orderscore{53.82}{+0.32}
& \orderscore{37.50}{+0.29}
& \orderscore{45.73}{-5.92}
& \orderscore{34.47}{-2.33} \\

MedDr
& \orderscore{61.44}{-0.52}
& \orderscore{45.56}{-1.32}
& \orderscore{70.38}{-0.32}
& \orderscore{51.47}{-1.04}
& \orderscore{51.10}{-1.11}
& \orderscore{37.20}{-0.93} \\

WSI-LLaVA
& \orderscore{62.50}{-3.26}
& \orderscore{54.08}{-8.29}
& \orderscore{57.96}{+0.64}
& \orderscore{56.75}{-6.36}
& \orderscore{40.76}{+4.36}
& \orderscore{36.10}{+2.31} \\

ALPaCA
& \orderscore{56.28}{-7.52}
& \orderscore{57.08}{-1.93}
& \orderscore{46.50}{-4.46}
& \orderscore{56.32}{-3.69}
& \orderscore{36.86}{+5.61}
& \orderscore{31.84}{+3.04} \\

MedGemma
& \orderscore{45.33}{-1.95}
& \orderscore{33.73}{-2.30}
& \orderscore{47.13}{-3.19}
& \orderscore{37.81}{-2.24}
& \orderscore{48.39}{-0.32}
& \orderscore{34.70}{+0.58} \\

PRISM2
& \orderscore{55.92}{-0.28}
& \orderscore{57.35}{-1.51}
& \orderscore{61.94}{-0.48}
& \orderscore{57.16}{-1.10}
& \orderscore{\underline{57.08}}{-0.09}
& \orderscore{43.23}{+0.51} \\

SlideChat
& \orderscore{\textbf{76.14}}{-6.25}
& \orderscore{\underline{62.16}}{-8.14}
& \orderscore{\underline{73.73}}{-0.79}
& \orderscore{\underline{66.85}}{-7.22}
& \orderscore{36.67}{+16.45}
& \orderscore{32.94}{+0.29} \\
\midrule

\rowcolor{oursblue}
\textbf{SlideJev}
& \orderscore{\underline{76.09}}{0.00}
& \orderscore{\textbf{66.26}}{0.00}
& \orderscore{\textbf{73.89}}{0.00}
& \orderscore{\textbf{69.53}}{0.00}
& \orderscore{\textbf{59.93}}{0.00}
& \orderscore{\textbf{44.61}}{0.00} \\
\bottomrule[0.8pt]
\end{tabularx}

\par\vspace{2pt}
{\footnotesize\raggedright
\textbf{Bold} and \underline{underlining} mark the best and
second-best four-order means, respectively.
Gray values: $\Delta=\mathrm{Mean}-\mathrm{Original}$ (pp).
\par}
\end{table*}

\begin{table*}[t]
\centering
\caption{\textbf{Results on BCNB.}
Each cell reports the four-order mean (\%); smaller gray
numbers show changes from the original-order result in
percentage points (pp).
Task columns and ACC report accuracy; BA denotes balanced
accuracy. The two HER2 tasks are combined.}
\label{tab:bcnb_main}

\small
\setlength{\tabcolsep}{2.5pt}
\renewcommand{\arraystretch}{1.10}

\begin{tabularx}{\textwidth}{
    @{}l*{8}{>{\centering\arraybackslash}X}@{}
}
\toprule[0.8pt]
\multirow{2}{*}{\textbf{Model}}
& \multicolumn{2}{c}{\textbf{Histology}}
& \multicolumn{3}{c}{\textbf{Biomarkers}}
& \multirow{2}{*}{
    \shortstack{\textbf{Molecular}\\\textbf{subtype}}
  }
& \multicolumn{2}{c}{\textbf{Overall}} \\
\cmidrule(lr){2-3}
\cmidrule(lr){4-6}
\cmidrule(lr){8-9}
& Subtype & Grade & ER & PR & HER2
& & \textbf{ACC} & \textbf{BA} \\
\midrule

\rowcolor{groupgray}
\multicolumn{9}{l}{\textbf{(a) General-purpose VLMs}} \\

Gemma-4-E4B-it
& \orderscore{87.04}{-0.08}
& \orderscore{52.46}{+0.16}
& \orderscore{20.10}{-0.08}
& \orderscore{27.58}{-0.49}
& \orderscore{\underline{45.93}}{-0.19}
& \orderscore{18.68}{+0.91}
& \orderscore{42.30}{+0.01}
& \orderscore{37.09}{+0.36} \\

Qwen3.6-27B
& \orderscore{22.25}{+0.25}
& \orderscore{30.12}{-2.35}
& \orderscore{48.26}{-2.38}
& \orderscore{36.38}{-3.95}
& \orderscore{40.55}{-2.00}
& \orderscore{12.98}{-0.15}
& \orderscore{33.08}{-1.80}
& \orderscore{37.83}{-2.55} \\

Qwen3.8-27B
& \orderscore{22.79}{-0.30}
& \orderscore{41.99}{-0.68}
& \orderscore{32.03}{+2.55}
& \orderscore{25.59}{+0.07}
& \orderscore{42.62}{-0.18}
& \orderscore{15.89}{0.00}
& \orderscore{31.75}{+0.20}
& \orderscore{38.00}{+0.40} \\

GPT-6.1-sol
& \orderscore{84.44}{+0.30}
& \orderscore{\textbf{58.82}}{-0.75}
& \orderscore{\underline{80.26}}{-0.05}
& \orderscore{\underline{74.21}}{+0.31}
& \orderscore{42.40}{-0.15}
& \orderscore{26.85}{-0.89}
& \orderscore{\underline{58.51}}{-0.19}
& \orderscore{39.95}{+0.02} \\
\midrule

\rowcolor{groupgray}
\multicolumn{9}{l}{\textbf{(b) Medical and pathology VLMs}} \\

Quilt-LLaVA
& \orderscore{49.98}{+0.62}
& \orderscore{13.72}{-0.61}
& \orderscore{49.05}{-2.86}
& \orderscore{47.15}{-8.30}
& \orderscore{30.44}{-13.65}
& \orderscore{25.00}{+0.42}
& \orderscore{35.50}{-5.51}
& \orderscore{37.30}{-3.75} \\

MedDr
& \orderscore{23.51}{-0.28}
& \orderscore{40.43}{0.00}
& \orderscore{19.52}{+0.13}
& \orderscore{25.39}{+0.07}
& \orderscore{44.11}{-0.42}
& \orderscore{12.02}{+0.37}
& \orderscore{29.67}{-0.07}
& \orderscore{38.15}{+0.36} \\

WSI-LLaVA
& \orderscore{28.47}{+0.32}
& \orderscore{36.90}{+2.07}
& \orderscore{50.42}{-2.76}
& \orderscore{48.85}{-7.77}
& \orderscore{35.45}{-14.03}
& \orderscore{\textbf{31.54}}{-0.74}
& \orderscore{38.21}{-5.40}
& \orderscore{38.77}{-4.61} \\

ALPaCA
& \orderscore{74.63}{-0.69}
& \orderscore{\underline{54.03}}{-0.95}
& \orderscore{51.27}{+0.54}
& \orderscore{56.13}{+5.39}
& \orderscore{38.59}{+3.61}
& \orderscore{24.80}{+0.32}
& \orderscore{48.20}{+1.74}
& \orderscore{38.79}{+1.37} \\

MedGemma
& \orderscore{85.43}{-0.10}
& \orderscore{32.45}{+1.65}
& \orderscore{59.84}{-2.16}
& \orderscore{47.91}{-5.18}
& \orderscore{31.54}{-5.42}
& \orderscore{11.57}{-0.08}
& \orderscore{43.09}{-2.45}
& \orderscore{33.27}{-2.44} \\

PRISM2
& \orderscore{\textbf{90.46}}{-0.03}
& \orderscore{23.49}{-1.37}
& \orderscore{29.36}{+0.17}
& \orderscore{28.61}{-0.14}
& \orderscore{\textbf{46.24}}{+0.42}
& \orderscore{19.97}{+0.23}
& \orderscore{40.86}{-0.02}
& \orderscore{\underline{41.20}}{+0.09} \\

SlideChat
& \orderscore{89.89}{+0.20}
& \orderscore{19.29}{-1.20}
& \orderscore{76.84}{-2.89}
& \orderscore{63.54}{-7.02}
& \orderscore{35.65}{-13.53}
& \orderscore{17.92}{-0.05}
& \orderscore{48.93}{-5.50}
& \orderscore{37.89}{-4.76} \\
\midrule

\rowcolor{oursblue}
\textbf{SlideJev}
& \orderscore{\underline{90.29}}{0.00}
& \orderscore{43.78}{0.00}
& \orderscore{\textbf{80.51}}{0.00}
& \orderscore{\textbf{75.17}}{0.00}
& \orderscore{45.87}{0.00}
& \orderscore{\underline{28.53}}{0.00}
& \orderscore{\textbf{58.86}}{0.00}
& \orderscore{\textbf{43.30}}{0.00} \\
\bottomrule[0.8pt]
\end{tabularx}

\par\vspace{2pt}
{\footnotesize\raggedright
\textbf{Bold} and \underline{underlining} mark the best and
second-best four-order means, respectively.
Gray values: $\Delta=\mathrm{Mean}-\mathrm{Original}$ (pp).
\par}
\end{table*}

\subsection{Jev Readout}
\label{sec:jev_readout}

This module converts parallel candidate verification
results into a unified answer distribution, enabling
direct selection through a common scoring rule.

For candidate input $x_i$, the original language-model head
evaluates fixed support and rejection suffixes $y^+$ and $y^-$,
with complete log-likelihoods
$\ell_i^\pm=\log P(y^\pm\mid x_i;C)$.
Each likelihood includes all suffix tokens, with each token
conditioned on the preceding suffix tokens.
Both branches reuse the visual updates computed from
the verification prompt, ensuring the same visual conditioning
for support and rejection.
The Jev score and candidate probability are
\begin{equation}
s_i=\ell_i^+-\ell_i^-,
\qquad
p_i=\frac{\exp(s_i)}{\sum_{j=1}^{K}\exp(s_j)}.
\label{eq:jev_readout}
\end{equation}
A positive score favors support over rejection.
Joint normalization makes the separate verification results
comparable across candidates, and we select the candidate
with the highest probability without decoding an option label.

\subsection{Image-grounded Supervision}
\label{sec:image_supervision}

We train the visual modules to improve answer correctness
and make candidate preferences reflect tissue evidence.
Both goals are optimized through the Jev scores used
for answer selection.

\vspace{1mm}
\noindent\textbf{Answer And Image-pair Objectives.}
Let $s_i(I)$ and $p_i(I)$ denote the candidate score and
probability evaluated on image $I$.
For correct candidate index $y$, question-answer supervision
uses $\mathcal{L}_{\mathrm{QA}}=-\log p_y(I)$.

To supervise the influence of image content, we additionally
use labelled image pairs $(I_A,I_B)$ sharing the same
question $Q$ and candidate set $C$, but having different
correct labels $y_A\neq y_B$.
Their cross-image preference change is
$\Delta=[s_{y_A}(I_A)-s_{y_B}(I_A)]
-[s_{y_A}(I_B)-s_{y_B}(I_B)]$.
A positive $\Delta$ indicates a stronger preference for
$y_A$ over $y_B$ on $I_A$ than on $I_B$.
The paired loss is
\begin{equation}
\mathcal{L}_{\mathrm{pair}}
=
-\frac12\log[p_{y_A}(I_A)p_{y_B}(I_B)]
+\beta\log(1+e^{\mu-\Delta}).
\label{eq:paired_supervision}
\end{equation}
The first term supervises the correct answer for each image.
The second encourages $\Delta$ to exceed the margin $\mu$,
so candidate preferences change with the visual evidence
while the question and candidates remain fixed.

\vspace{1mm}
\noindent\textbf{Joint Optimization.}
The two objectives jointly train the visual modules:
\begin{equation}
\mathcal{L}
=
\mathcal{L}_{\mathrm{QA}}
+\lambda\mathcal{L}_{\mathrm{pair}},
\label{eq:training_objective}
\end{equation}
where $\mu>0$ is the preference-change margin,
$\beta$ weights the preference-change term in the paired
loss, and $\lambda$ controls paired supervision.
Each loss is averaged over its training examples.
With the pretrained backbone frozen, gradients through
Jev scores train the added visual modules, linking parallel
answer selection to image-dependent visual updates.
\section{Experiments}
\label{sec:experiments}

\subsection{Experimental Setup}
\label{sec:experimental_setup}

\vspace{1mm}
\noindent\textbf{Datasets And Tasks.}
We evaluate SlideJev on SlideBench-VQA~\cite{chen2025slidechat} for whole-slide multiple-choice question answering and assess transfer on BCNB~\cite{xu2021predicting} for whole-slide question answering and on BRACS~\cite{brancati2022bracs} for diagnostic classification.
Training uses multiple-choice examples from SlideChat's instruction-tuning dataset, SlideInstruction~\cite{chen2025slidechat}, supplementary samples from GTEx~\cite{gtex2020gtex}, and image pairs constructed from report-derived labels.
Dataset details are provided in \textbf{Appendix~\textcolor{cvprblue}{A}}.

\vspace{1mm}
\noindent\textbf{Implementation Details.}
We build SlideJev on SlideChat~\cite{chen2025slidechat} using CONCH~\cite{lu2024visual} patch features.
We jointly optimize the proposed evidence reading,
candidate comparison, and gated visual injection modules,
with the pretrained SlideChat backbone kept fixed.
For question-answer training, we set the preference margin
to $\mu=1$, the visual difference weight within the paired
loss to $\beta=0.2$, and the overall paired-supervision
weight to $\lambda=0.5$.
Additional details are provided in \textbf{Appendix~\textcolor{cvprblue}{B}}.

\vspace{1mm}
\noindent\textbf{Evaluation Metrics.}
We report accuracy (ACC) on all benchmarks and balanced
accuracy (BA) on BCNB and BRACS, whose tasks have predefined
semantic classes.
Option-order robustness is assessed under four orders:
original, reversed, cyclically shifted, and fixed random.
Answer consistency is the fraction of examples for which
all four predictions are valid and select the same candidate
after mapping back to the original order.
We assess image dependence through correct-image ACC gains
over mismatched-image and image-free inputs under the
original option order.

\subsection{Comparison With State-of-the-art Methods}
\label{sec:main_results}

\vspace{1mm}
\noindent\textbf{Comparison Methods.}
We compare SlideJev with general-purpose VLMs, including
Gemma-4-E4B-it~\cite{team2026gemma},
Qwen3.6-27B~\cite{qwen2026qwen36},
Qwen3.8-27B~\cite{qwen2026qwen38},
and GPT-6.1-sol~\cite{openai2026gpt61sol},
and with medical and pathology VLMs, including
Quilt-LLaVA~\cite{seyfioglu2024quilt},
MedDr~\cite{he2024meddr},
WSI-LLaVA~\cite{liang2025wsi},
ALPaCA~\cite{gao2025alpaca},
MedGemma~\cite{sellergren2025medgemma},
PRISM2~\cite{vorontsov2026end},
and SlideChat~\cite{chen2025slidechat}.
Model-specific visual input protocols and inference settings
are provided in the supplementary material.
Baseline settings are provided in \textbf{Appendix~\textcolor{cvprblue}{C}}.

\vspace{1mm}
\noindent\textbf{Main Results.}
On SlideBench-VQA, SlideJev achieves the highest mean overall ACC across four option orders, as shown in Table~\ref{tab:tcga_bracs}.
The baseline that leads under the original order shows a substantial accuracy drop after reordering, while SlideJev maintains its performance.
BRACS-3 further evaluates transfer to diagnostic classification under class imbalance.
SlideJev leads in both ACC and BA, showing that its advantage remains when all classes receive equal weight in evaluation.
BCNB covers histological assessment, biomarker prediction, and molecular subtyping, with different baselines performing well on different tasks.
Table~\ref{tab:bcnb_main} shows that SlideJev achieves the highest four-order mean overall ACC and mean BA across these tasks.
It also retains its original-order performance, whereas several baselines show substantial drops in overall accuracy after reordering.
Across the three benchmarks, SlideJev's accuracy advantage under four-order evaluation extends to both question answering and diagnostic classification.
Additional results are provided in \textbf{Appendix~\textcolor{cvprblue}{D}}.

\begin{figure}[!t]
    \centering
    \includegraphics[width=\columnwidth]
    {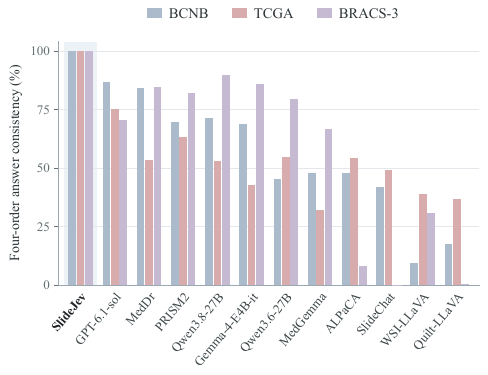}
    \caption{\textbf{Answer consistency across four option orders.}
SlideJev maintains 100\% consistency across all three
benchmarks, while the compared baselines remain sensitive
to option order.}
    \label{fig:option_order_consistency}
\end{figure}

\vspace{1mm}
\noindent\textbf{Option-order Consistency.}
We evaluate answer consistency across four option orders.
As shown in Figure~\ref{fig:option_order_consistency}, SlideJev maintains perfect consistency across all three benchmarks, while baseline predictions remain sensitive to option order.
Parallel inputs preserve each candidate's verification context, enabling consistent answer selection through Jev scoring.

\begin{figure}[!ht]
    \centering
    \includegraphics[width=\columnwidth]
    {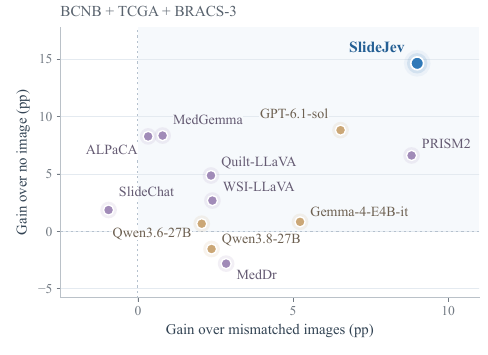}
    \caption{\textbf{Correct-image accuracy gains.}
    Horizontal and vertical axes show original-order ACC gains
    (pp) over mismatched and image-free inputs, respectively,
    averaged equally across BCNB, SlideBench-VQA, and BRACS-3.}
    \label{fig:visual_gains}
\end{figure}

\vspace{1mm}
\noindent\textbf{Effect of Visual Evidence.}
We use mismatched slides from examples of the same task
type but with different ground-truth answers.
SlideJev achieves the largest correct-image gains over
mismatched and image-free inputs
(Figure~\ref{fig:visual_gains}), indicating stronger
reliance on slide-specific evidence.

\begin{table}[!ht]
\centering
\caption{\textbf{Ablation study of SlideJev.}
Results are averaged equally across the full BCNB, TCGA,
and BRACS-3 evaluation cohorts.
ACC: four-order mean accuracy (\%); gray values:
changes from original-order ACC (pp).
Cons.: answer consistency (\%).
Image gains: original-order ACC gains (pp) over
mismatched (Wrong) and image-free (No image) inputs.}
\label{tab:ablation}

\small
\setlength{\tabcolsep}{2pt}
\renewcommand{\arraystretch}{1.10}

\begin{tabularx}{\columnwidth}{
    @{}lc*{3}{>{\centering\arraybackslash}X}@{}
}
\toprule[0.8pt]
\multirow{2}{*}{\textbf{Variant}}
& \multirow{2}{*}{\textbf{ACC}}
& \multirow{2}{*}{\textbf{Cons.}}
& \multicolumn{2}{c}{\textbf{Image gain}} \\
\cmidrule(lr){4-5}
& & & Wrong & No image \\
\midrule

w/o evidence module
& \orderscore{47.95}{0.00}
& 100.00
& $-0.02$
& $-0.17$ \\

w/o comparison
& \orderscore{55.90}{0.00}
& 100.00
& +4.87
& +7.78 \\

w/o parallel inputs
& \orderscore{55.24}{-5.15}
& 33.26
& +2.37
& +14.44 \\

w/o paired supervision
& \orderscore{53.75}{0.00}
& 100.00
& $-0.33$
& +5.60 \\

Generative readout
& \orderscore{46.88}{0.00}
& 99.88
& +1.47
& +12.82 \\
\midrule

\rowcolor{oursblue}
\textbf{Full SlideJev}
& \orderscore{\textbf{62.77}}{0.00}
& \textbf{100.00}
& \textbf{+9.00}
& \textbf{+14.65} \\
\bottomrule[0.8pt]
\end{tabularx}
\end{table}
\subsection{Ablation Study}
\label{sec:ablation}

\vspace{1mm}
\noindent\textbf{Component Ablations.}
\Cref{tab:ablation} evaluates the evidence module,
candidate comparison, parallel inputs, paired supervision,
and answer readout.
Removing parallel inputs replaces independent verification
with joint prompting and constrained option-letter generation,
reducing both ACC and answer consistency.
Generative readout retains parallel inputs and the evidence
module but replaces Jev scoring with numeric support generation,
preserving high consistency at lower ACC.
Removing the evidence module, candidate comparison,
or paired supervision also reduces ACC and correct-image gains.

\begin{figure}[!ht]
    \centering
    \includegraphics[width=\columnwidth]
    {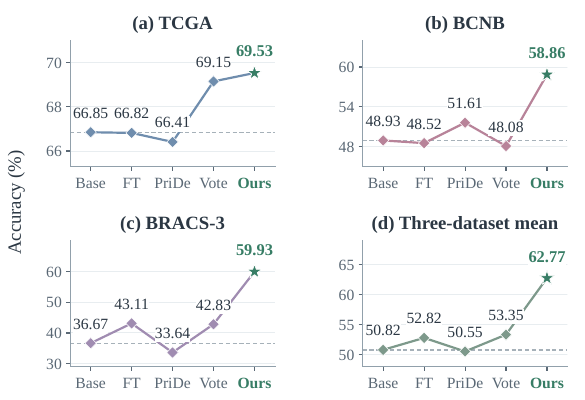}
    \caption{\textbf{Additional SlideChat baseline comparisons.}
    FT denotes same-data fine-tuning.
    Non-voting methods report four-order mean ACC;
    Vote aggregates predictions across the four orders.
    Panel (d) reports the equal-weight mean across datasets.
    Dashed lines indicate SlideChat; vertical scales vary by panel.}
    \label{fig:additional_baselines}
\end{figure}

\vspace{1mm}
\noindent\textbf{Additional Baseline Comparisons.}
SlideJev achieves higher mean accuracy than alternative strategies for improving SlideChat.
We apply same-data fine-tuning, PriDe~\cite{zheng2023large}, and four-order voting independently to SlideChat, examining the benefits of additional training, bias calibration, and prediction aggregation across option orders, respectively.
\Cref{fig:additional_baselines} summarizes the comparison across the three datasets.
Further results appear in \textbf{Appendix~\textcolor{cvprblue}{D}}.

\begin{figure*}[t]
    \centering
    \includegraphics[width=\textwidth]{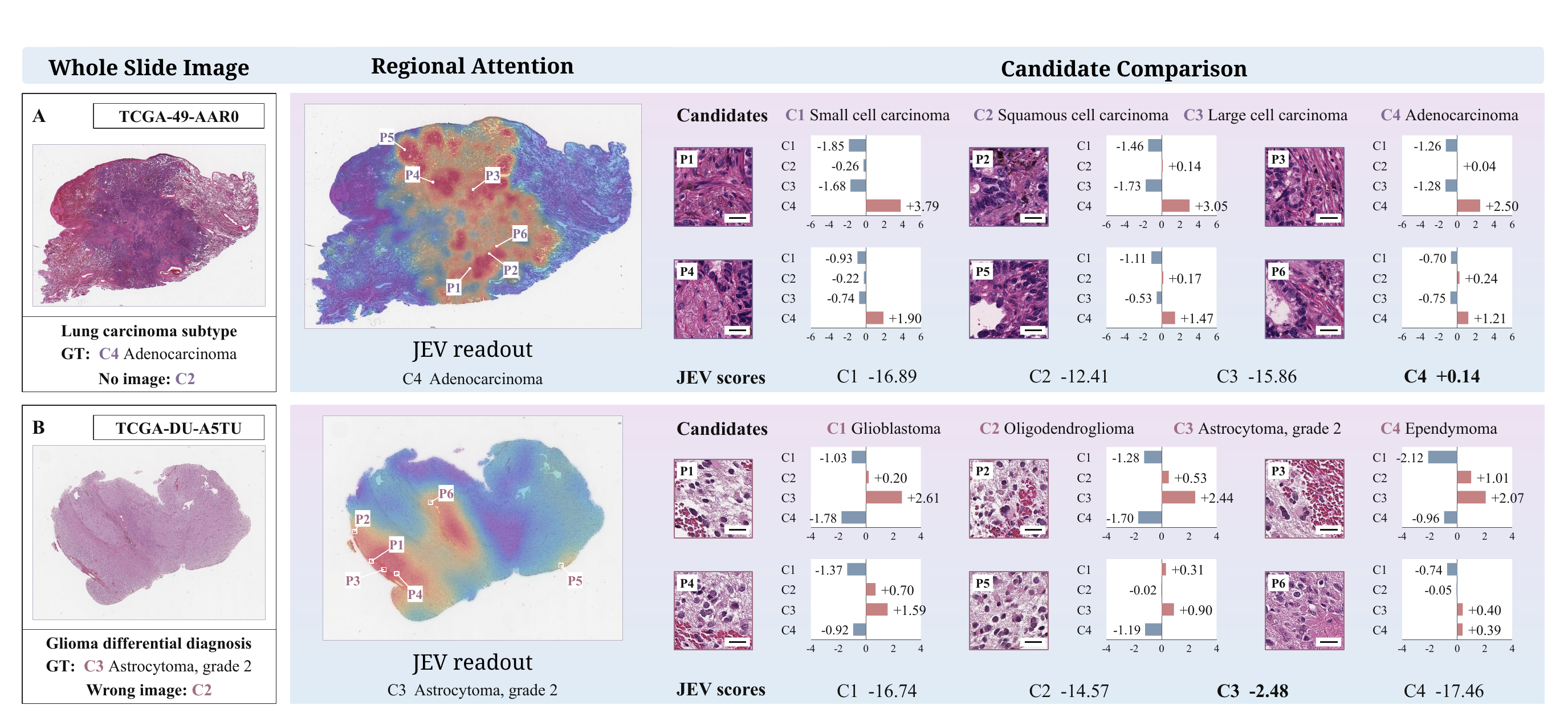}
    \caption{\textbf{Case studies of regional attention and Jev readout.}
    (A) Lung carcinoma subtyping. (B) Glioma differential diagnosis.
    Heatmaps show layer-averaged attention for the predicted candidate,
    with six patches ranked by attention.
    Bars show percentage deviations from each patch's cross-candidate
    mean attention; red and blue indicate above- and below-mean values.
    Jev scores appear below each case.
    Cases A and B include image-free and mismatched-image predictions,
    respectively.}
    \label{fig:case_study}
\end{figure*}
\section{Interpretability}
\label{sec:case_study}

\Cref{fig:case_study} links regional attention, patch-level
candidate comparison, and Jev readout in a lung carcinoma
subtype case and a glioma differential diagnosis case.
Heatmaps show layer-averaged reader attention for the
predicted candidate, while enlarged patches reveal the
corresponding tissue morphology.
For each selected patch, bars show attention deviations
from the cross-candidate mean, highlighting differences
in candidate responses to the same tissue evidence.

In the lung carcinoma case, the image-free input favors
squamous cell carcinoma.
With the correct WSI, adenocarcinoma receives the highest
Jev score and is correctly selected.
In the glioma case, a mismatched image favors
oligodendroglioma, whereas the correct WSI leads to
astrocytoma, grade 2, matching the ground truth.
These cases show how candidate preferences change
with visual input, while attention maps and comparisons
reveal the tissue evidence associated with these decisions.
More visualizations are provided in
\textbf{Appendix~\textcolor{cvprblue}{E}}.
\section{Related Work}
\label{sec:related_work}

\vspace{1mm}
\noindent\textbf{Pathology Vision-language Models.}
PathVQA introduced question answering over pathology
images~\cite{he2020pathvqa}.
Image-level models, including MedDr, MedGemma, and
Quilt-LLaVA, adapt visual-language learning to medical
and pathology images~\cite{he2024meddr,sellergren2025medgemma,
seyfioglu2024quilt}.
Related efforts strengthen pathology-specific image-text
alignment and instruction learning~\cite{lu2024multimodal,
sun2024pathasst,sun2025pathgen}.
However, thumbnails and selected image fields provide
limited tissue detail or coverage for whole-slide questions.
Pathology encoders and slide representation models provide
foundations for broader tissue understanding through
self-supervised and multimodal learning~\cite{lu2024visual,
xu2024whole,zhang2026care,shaikovski2024prism,ding2025multimodal}.
Building on whole-slide representations, SlideChat supports
gigapixel WSI understanding through instruction
learning~\cite{chen2025slidechat}, while WSI-LLaVA emphasizes
morphological findings to explain its answers~\cite{liang2025wsi}.
ALPaCA combines question-dependent visual-text interaction
with tissue prototypes~\cite{gao2025alpaca}, and PRISM2
uses clinical dialogue supervision for slide-level
inference~\cite{vorontsov2026end}.
SlideJev follows this paradigm while focusing on
stable candidate verification and discriminative visual
evidence use.

\vspace{1mm}
\noindent\textbf{Reliable Candidate Decisions.}
Multiple-choice predictions can depend on option
representation and order~\cite{robinson2022leveraging,
zheng2023large,pezeshkpour2024large}.
Calibration and PriDe's option-prior correction address
answer biases~\cite{zhao2021calibrate,zheng2023large},
while VQA studies seek to reduce language shortcuts
and strengthen visual grounding~\cite{goyal2017making,
agrawal2018don,leng2024mitigating}.
Jev-style interfaces expose probabilities over declared
answer spaces, and LLM-as-Jev obtains such judgments
through language-model probability readout~\cite{typesafe2026jev,
li2026llmasjev}.
Recent evaluations examine these judgments across general
and medical tasks~\cite{deusser2026evaluating,madrid2026jev}.
SlideJev connects these directions through parallel
candidate verification and candidate-conditioned visual
updates.
Shared Jev scores support answer selection and supervision
that encourages candidate preferences to reflect tissue evidence.
\section{Conclusion}

We presented SlideJev, a whole-slide pathology VQA framework
combining parallel candidate verification and gated visual injection.
Independent, identically formatted prompts without
option-position labels preserve verification contexts
under reordering.
A shared Jev readout connects answer selection with
visual evidence learning.
Question-answer and paired-image supervision train the
visual modules through these scores, with the backbone frozen.
Across SlideBench-VQA, BCNB, and BRACS-3, SlideJev achieves
leading four-order mean overall accuracy among the compared
methods, consistent candidate selection across option orders,
and stronger dependence on the correct slide.
By connecting order-stable candidate verification with
tissue evidence learning through a shared Jev interface,
SlideJev supports reliable pathological decisions grounded
in relevant, slide-specific morphological findings
that inform both diagnostic and phenotypic assessment.

{
    \small
    \bibliographystyle{slidejev_refs}
    \bibliography{main}
}

\clearpage

\end{document}